\documentclass{article}

\usepackage[preprint]{corl_2026} 

\makeatletter
\if@preprinttype
\renewcommand{\@noticestring}{Preprint.}
\else
\renewcommand{\@noticestring}{%
  \@conferenceordinal\/ Conference on Robot Learning (CoRL \@conferenceyear), \@conferencelocation.%
}
\fi
\makeatother

\usepackage{graphicx}
\usepackage{wrapfig}
\usepackage{siunitx} 
\usepackage{amssymb}
\usepackage{longtable}
\usepackage{subcaption}

\newcommand{\vspaceafter}{\vspace{-0mm}}
\definecolor{unfoldyellow}{RGB}{255,250,199}
\definecolor{unfoldgreen}{RGB}{216,255,213}
\definecolor{unfoldblue}{RGB}{202,252,255}

\title{FoldNet++: a Large-Scale Synthetic Dataset for Robotic T-Shirt Folding and Unfolding}

\author{
Yuxing Chen$^{1,2,*}$ \And Zhiyuan Wei$^{1,2,*}$ \And Bowen Xiao$^{1,2}$ \And Zhizheng Zhang$^{2,\dagger}$ \And He Wang$^{1,2,\dagger}$ \\
~ \\
$^{1}$ CFCS, Peking University, China \\
$^{2}$ Galbot, China \\
$^{*}$ Equal Contribution, $^{\dagger}$ Corresponding Author
}

\begin{document}
\maketitle


\begin{abstract}
Due to the highly deformable nature of garments, training a generalizable policy for robotic T-shirt folding and unfolding remains a significant challenge. In this work, we present a large-scale synthetic dataset for robotic T-shirt folding and unfolding, covering 6 robotic embodiments, 1K T-shirts, 1K environmental assets, and 120K episodes with rich annotations, which can be used to train a wide range of manipulation policies. We first follow the FoldNet pipeline to generate a large-scale dataset of physically simulatable T-shirts with diverse appearances and annotated semantic keypoints. Based on these semantic keypoints, we then generate manipulation demonstrations for different robotic embodiments through a unified rule-based framework. We use these demonstrations to train visuomotor policies, and experimental results demonstrate that models trained solely on our synthetic data can achieve over 90\% end-to-end task success rates when directly deployed to unseen real-world environments and previously unseen T-shirts from arbitrary initial configurations. Project URL: \url{https://pku-epic.github.io/FoldNetXX/}.
\end{abstract}



\section{Introduction}
\label{sec:intro}

Garment folding and unfolding have long been regarded as fundamental yet highly challenging problems in deformable object manipulation~\cite{Longhini2025unfoldingtheliterature}. The primary difficulty stems from the highly deformable nature of garments, which requires a robust policy capable of handling a wide variety of cloth configurations and dynamics. Moreover, garment manipulation is inherently a long-horizon task consisting of multiple sequential subtasks, further increasing the complexity of planning and control.

In recent years, data-driven learning approaches have emerged as a promising solution for garment manipulation~\cite{intelligence2025pi06vlalearnsexperience,zheng2026xvla,yu2026chi0resourceawarerobustmanipulation,zheng2026egoscalescalingdexterousmanipulation,chen2025foldnetlearninggeneralizableclosedloop}. These methods rely on large-scale robotic demonstration data and leverage imitation learning to enable robots to predict appropriate manipulation actions under diverse observations. Given sufficiently diverse and high-quality demonstrations, the learned policies can generalize across different garment states and manipulation scenarios. 

However, existing approaches still suffer from several important limitations. First, collecting real-world robotic manipulation data is extremely expensive and time-consuming. Most prior works~\cite{intelligence2025pi06vlalearnsexperience,zheng2026xvla,yu2026chi0resourceawarerobustmanipulation,zheng2026egoscalescalingdexterousmanipulation} rely heavily on real-world demonstrations for training, but such datasets are difficult to scale and often lack sufficient diversity in environments, robot embodiments, and garment appearances. Moreover, changes in robot platforms, deployment environments, or target garments typically require collecting new datasets.

Leveraging synthetic data generated in simulation provides a scalable and cost-effective approach for training garment manipulation policies~\cite{chen2025foldnetlearninggeneralizableclosedloop,tian2025interndataa1pioneeringhighfidelitysynthetic,deng2025graspvla,zhao2026simreal,cai2026internvlaa1unifyingunderstandinggeneration}. Compared with real-world data collection, simulation environments enable extensive randomization over environments, objects, and robot embodiments, thereby offering significantly greater potential for policy generalization. However, existing simulation-based pipelines often rely on simplified task settings. In particular, prior synthetic data generation approaches~\cite{chen2025foldnetlearninggeneralizableclosedloop,tian2025interndataa1pioneeringhighfidelitysynthetic} typically assume that garments are already flattened on a table before the folding process begins. As a result, these methods avoid the more challenging problem of manipulating garments from arbitrary and highly crumpled initial configurations. Developing robust sim-to-real policies capable of handling such complex garment states remains a largely open problem.

\begin{figure}[t]
\centering
\includegraphics[width=1.0\linewidth]{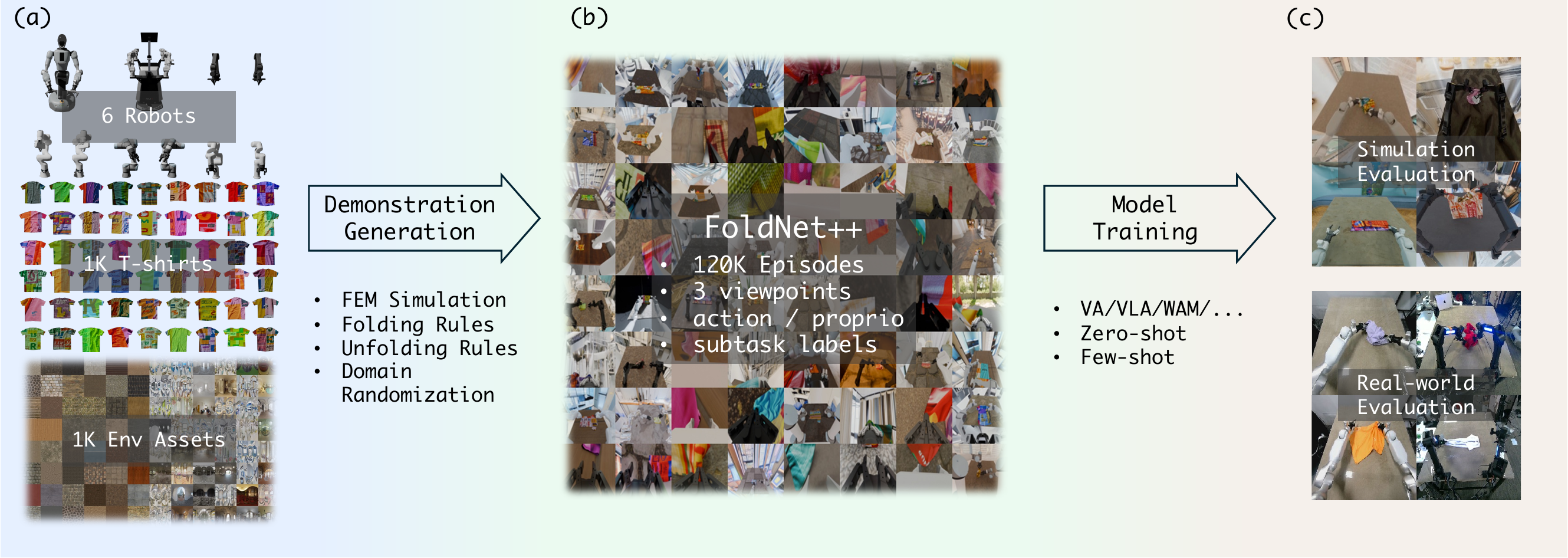}
\caption{
Overview of FoldNet++. (a) Our dataset includes 6 robots, 1K T-shirts, and 1K table textures and HDRI environment maps. (b) Using these assets, we import them into an FEM-based physics simulator. Combined with predefined manipulation rules and domain randomization, we generate 120K episodes containing multi-view observations from three camera viewpoints, robot actions, and subtask annotations. (c) Leveraging the FoldNet++ dataset, we train different categories of models (VA, VLA, and WAM) and evaluate them across multiple embodiments in both simulated and real-world environments under zero-shot and few-shot settings.
}
\label{fig:teaser}
\vspaceafter
\end{figure}

In this work, we introduce FoldNet++, a large-scale synthetic dataset for T-shirt folding and unfolding. The dataset contains 6 different robot embodiments, 1K T-shirts, 1K diverse environments, and 120K episodes with detailed annotations. FoldNet++ can be directly used to train manipulation models capable of folding T-shirts from arbitrary initial configurations, while its unfolding component can be employed to flatten T-shirts, facilitating subsequent downstream tasks such as hanging or ironing. 

To generate this dataset, we first adopt the method proposed in~\cite{chen2025foldnetlearninggeneralizableclosedloop} to generate T-shirt meshes with semantic keypoint annotations. We then design a rule-based manipulation framework that leverages these semantic keypoints to perform garment unfolding and folding from arbitrary initial configurations across different robot platforms.

Using FoldNet++, we conduct extensive experiments to validate the effectiveness of the dataset. We evaluate multiple policy architectures, including vision-action models (VA)~\cite{chi2023diffusionpolicy}, vision-language-action models (VLA)~\cite{black2026pi0visionlanguageactionflowmodel}, and world action models (WAM)~\cite{zhu2025uwm}, in both simulation and real-world settings under zero-shot and few-shot conditions. Experiments are conducted on both in-distribution and out-of-distribution robot embodiments. Real-world evaluations demonstrate that policies trained solely on our synthetic data can achieve over 90\% end-to-end task success rates from arbitrary initial garment states, reaching performance comparable to prior approaches trained with large-scale real-world datasets.

Compared to FoldNet, FoldNet++ extends prior work in three key aspects: (1) we address the substantially more challenging problem of unfolding-to-folding manipulation from arbitrary, highly crumpled garment states; (2) we construct a large-scale cross-embodiment manipulation dataset spanning 6 robot embodiments; and (3) we conduct comprehensive quantitative evaluations, including cross-embodiment generalization, comparisons across different categories of manipulation policies, and real-data fine-tuning.
	

\section{Related Works}
\label{sec:related_works}

\subsection{Garment Manipulation}

Traditional approaches decompose garment manipulation into a sequence of action primitives~\cite{canberk2022clothfunnelscanonicalizedalignmentmultipurpose,xue2023unifolding,Wu_2024_CVPR,zhuang2025flatnfoldadiverse,li2025sktintegratingstateaware,oriol2025bifold,sunil2025reactiveinairclothingmanipulation}. These methods typically benefit from efficient training and stable control performance. However, operating in relatively low-dimensional action spaces, they often struggle to capture the fine-grained manipulations required for complex garment interactions. Another class of methods explicitly models garment dynamics to plan manipulation actions~\cite{Huang-RSS-22,tian2025diffusion,chen2025graphgarment,chen2025metafold}. While such approaches provide stronger structural priors, they remain sensitive to modeling inaccuracies and simulation errors.

Recently, there has been growing interest in training end-to-end garment manipulation policies via behavior cloning on large-scale real-world datasets~\cite{black2025pi05,zheng2026xvla,yu2026chi0resourceawarerobustmanipulation}. Some works further incorporate world models~\cite{gigabrainteam2026gigabrain05mvlalearnsworld,nvidia2025cosmosworldfoundationmodel} or reinforcement learning~\cite{intelligence2025pi06vlalearnsexperience,pan2026sopscalableonlineposttraining} to improve policy performance and robustness. Despite promising results, these approaches typically require expensive real-world data collection and often exhibit limited generalization to unseen environments and embodiments.

In contrast, by training end-to-end policies on large-scale synthetic data, our approach achieves strong generalization to unseen environments while retaining the ability to perform fine-grained, long-horizon garment manipulation.

\subsection{Garment Simulation Environment}

There has been substantial progress in the development of simulation environments~\cite{lu2024garmentlab,li2025lehome} and synthetic asset datasets~\cite{Zhou_2023_ICCV,chen2025foldnetlearninggeneralizableclosedloop,tian2025interndataa1pioneeringhighfidelitysynthetic} for garment manipulation. Several prior works~\cite{chen2025foldnetlearninggeneralizableclosedloop,tian2025interndataa1pioneeringhighfidelitysynthetic} further generate synthetic manipulation datasets to train learning-based policies. Despite these advances, the simulators adopted in previous data generation pipelines still exhibit a noticeable sim-to-real gap, particularly when garments undergo large and complex deformations. Moreover, the folding and unfolding procedures used in prior works are often overly simplified: garment folding is typically implemented by folding the sleeves followed by a single halving operation~\cite{chen2025foldnetlearninggeneralizableclosedloop,tian2025interndataa1pioneeringhighfidelitysynthetic}, which does not adequately capture the complexity of real-world folding behaviors. In addition, high-quality synthetic demonstrations for garment unfolding remain largely unavailable, primarily due to the difficulty of manipulating heavily crumpled garments from arbitrary initial configurations.

In this work, we leverage a high-fidelity physics simulator that preserves accurate physical behavior even under severe garment deformations. Building upon this simulator, we further introduce a scalable rule-based pipeline for synthesizing high-quality unfolding demonstrations, enabling learning-based methods to handle more realistic garment manipulation scenarios.


\section{FoldNet++ Dataset}
\label{sec:method}

Our demonstration generation pipeline first adopts the method proposed in~\cite{chen2025foldnetlearninggeneralizableclosedloop} to generate a large collection of T-shirts annotated with semantic keypoints, where each keypoint (e.g., left shoulder, bottom-left corner) corresponds to a specific vertex index. These T-shirts are then imported into a physics simulator for dynamic simulation. Owing to the keypoint annotations provided in~\cite{chen2025foldnetlearninggeneralizableclosedloop}, we can access the 3D coordinates of each keypoint at any time during simulation. Based on these keypoint coordinates, and combined with the robot’s inverse kinematics (IK), we design rule-based robot motions. Sec.~\ref{ssec:system_setup} describes the robotic system. Sec.~\ref{ssec:folding_rules} describes the garment folding rules, while Sec.~\ref{ssec:unfolding_rules} presents the garment unfolding rules. Sec.~\ref{ssec:simulation_environment} introduces the simulation environment. 

\subsection{Robot Setup}
\label{ssec:system_setup}

\begin{wrapfigure}[11]{r}{0.60\textwidth}
\centering
\vspace{-15mm}
\includegraphics[width=1.0\linewidth]{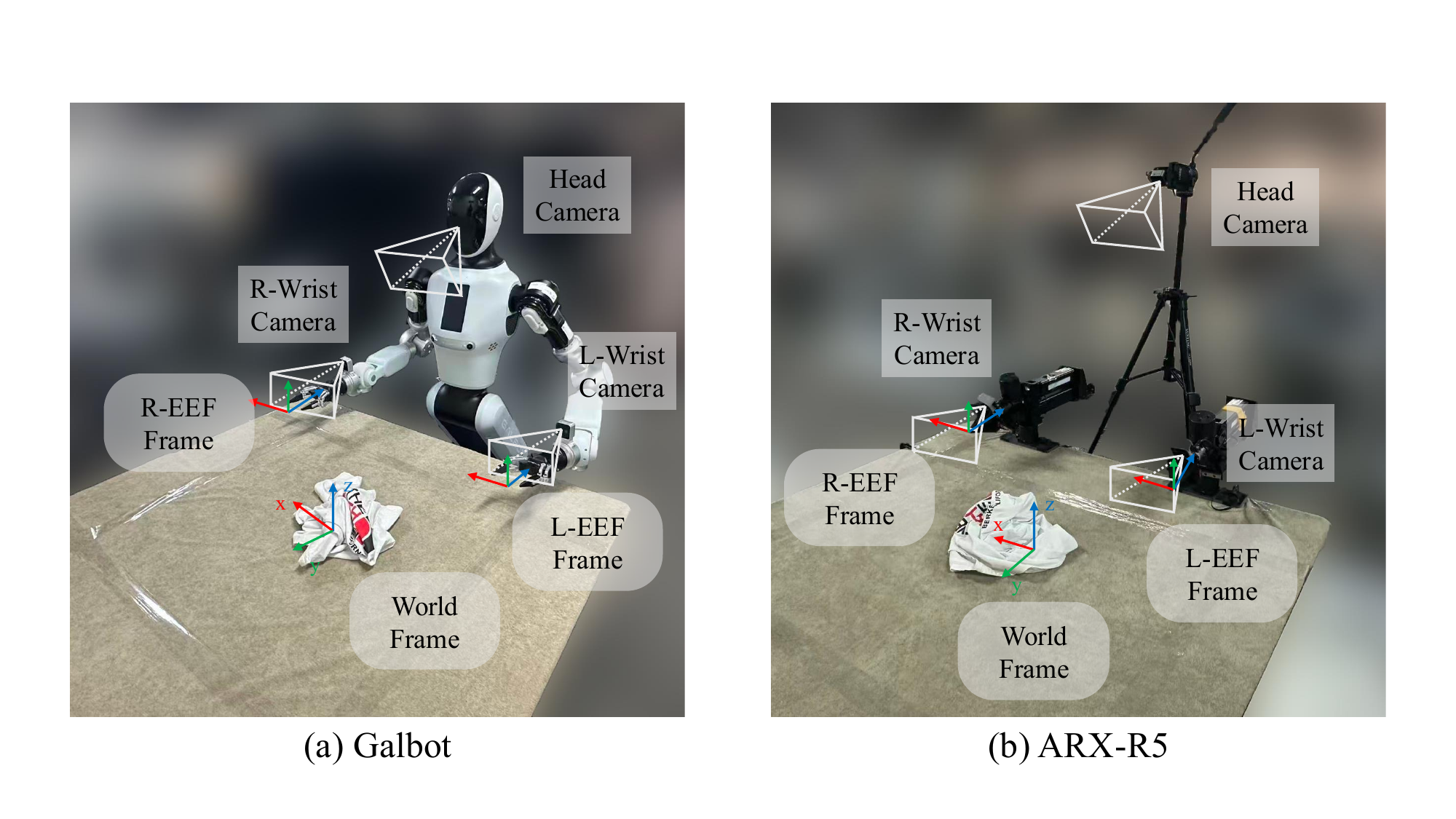}
\caption{Robot setup.}
\label{fig:setup}
\end{wrapfigure}

The setup of our robotic system is illustrated in Fig.~\ref{fig:setup}. The robot takes observation images from three viewpoints (head-mounted, left wrist-mounted, and right wrist-mounted cameras) as input and outputs the 6D pose of the end-effector (EEF) in the world frame. For different robotic embodiments, the EEF frames are aligned. We also align the manipulation spaces across robots by manually defining the 6D pose of each robot base relative to the world frame, enabling all robots to operate near the origin of the world frame. The tabletop is approximately located on the plane $z=0$.

The garment is initially placed near the origin of the world frame. The robot is required to start from arbitrary initial garment configurations, first flatten the garment, and then fold it into a neat configuration. In simulation, we randomly select a point on the garment, suspend it at $(0, 0, \SI{0.6}{\meter})$, and then released, allowing the garment to fall freely. In the real world, we follow a similar procedure by having the robot grasp a random point on the garment and release it from a height.

\subsection{Folding Rules}
\label{ssec:folding_rules}

Assume that the garment has already been flattened, with the collar oriented either to the left or to the right, roughly aligned with the world frame, and with its center located near the origin of the world coordinate system. In Fig.~\ref{fig:fold_vis_unfold_vis}a, we illustrate the complete folding pipeline using the case where the collar faces right as an example.

Our garment folding strategy follows a linear sequence of operations: folding the half of the garment closer to the robot toward the garment’s midline (Fold1), dragging the garment toward the robot (Drag1), folding the half of the garment farther from the robot toward the midline (Fold2), dragging the garment away from the robot (Drag2), folding the garment from left to right (Fold3), flipping the garment (Flip), and finally moving it to the right (Drag3). These subtasks are annotated in the top-left corner of each subfigure in Fig.~\ref{fig:fold_vis_unfold_vis}a.

If the collar faces left, the corresponding sign conventions in the rules are adjusted accordingly. Alternative folding styles can be implemented within our rule-based framework by adding or removing steps as needed; different folding variants are beyond the scope of this work.

\subsection{Unfolding Rules}
\label{ssec:unfolding_rules}

\begin{figure}[t]
\centering
\includegraphics[width=1.0\linewidth]{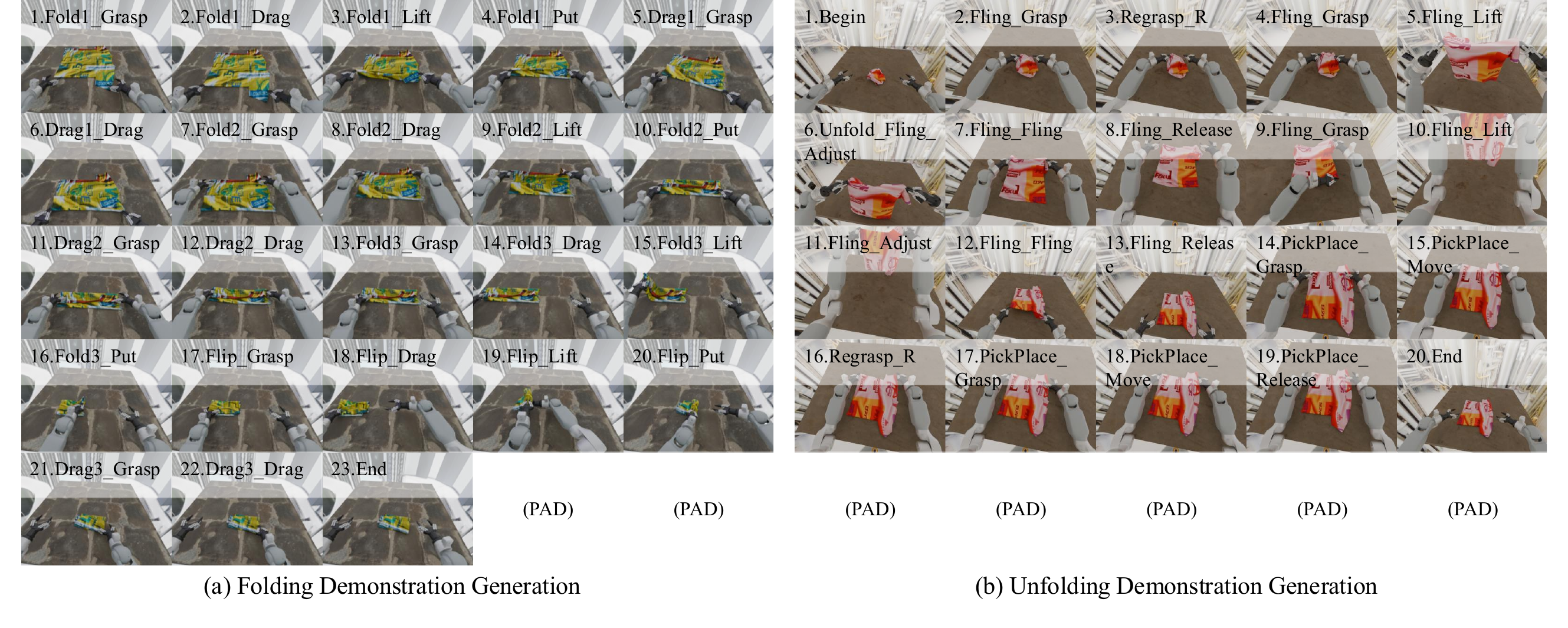}
\caption{
Head-view visualization of demonstration generation in simulation. (a) Example demonstration generated by the folding rules. (b) Example demonstration generated by the unfolding rules. 
}
\label{fig:fold_vis_unfold_vis}
\vspaceafter
\end{figure}

A sequence of head-view images illustrating the unfolding process from a random initial state is shown in Fig.~\ref{fig:fold_vis_unfold_vis}b. In general, our rule-based framework first employs a fling operation to flatten the garment, followed by pick-and-place actions to adjust its orientation and position~\cite{canberk2022clothfunnelscanonicalizedalignmentmultipurpose,xue2023unifolding}. During the fling stage, we first compute the manipulability of each vertex. We then select the pair of manipulable vertices that is most effective for spreading the garment and execute the fling action.

Once the garment becomes sufficiently flattened, we switch to a pick-and-place strategy, dragging the garment toward the center of the table while aligning the collar either to the left or to the right, approximately aligned with the world coordinate frame. Detailed implementation procedures are provided in the appendix.

\subsection{Simulation Environment}
\label{ssec:simulation_environment}

In our experiments, we use Style3D~\cite{style3d} as the physics engine for cloth simulation and Blender~\cite{blender} as the rendering engine. Style3D employs a finite element method (FEM) based on elastic mechanics, providing higher physical accuracy and efficient simulation for large deformations. For garment grasping, we adopt an attachment-based approach, where vertices within a certain distance of the EEF are attached to the gripper once it closes. In the simulation environment, the time step is \SI{0.2}{\second}, and a single demonstration trajectory, including both unfolding and folding phases, consists of approximately 300 steps.

For environment assets, we use PolyHaven~\cite{polyhaven}, which includes 268 table materials and 214 indoor HDRIs. Additionally, we generate 762 indoor HDRIs using the methods in~\cite{tang2023MVDiffusion,Wang_2025_CVPR}. We generate 1,000 T-shirt instances using the method in~\cite{chen2025foldnetlearninggeneralizableclosedloop}. For table textures, HDRIs, and garment instances, 80\% of the data is used for training, while the remaining 20\% is reserved for simulation testing. We collect data across six embodiments: Galbot, AgileX, dual-arm ARX-R5, dual-arm UR5e equipped with Robotiq grippers, dual-arm Franka, and dual-arm xArm equipped with Robotiq grippers. 

We apply domain randomization to improve sim-to-real transfer performance. The primary randomization factors include camera intrinsic and extrinsic parameters, robot base poses, and the physical properties of the cloth. In addition, when the robot gripper is open, we randomly select 5\% of the timesteps (each timestep corresponds to \SI{0.2}{\second}) to apply perturbations to the states of both the garment and the robot. Since no interaction occurs between the robot and the garment at these moments, such perturbations do not introduce artifacts. These perturbations are intended to encourage the model to rely more on visual observations rather than robot proprioceptive states, which we empirically observe to improve robustness in practice.


\section{Experimental Results}
\label{sec:experimental_results}

In this section, we present the experimental setup in Sec.~\ref{ssec:experimental_setup}. The subsequent experiments are primarily intended to answer the following questions: (1) How does the model perform across different embodiments? (2) How effective is the sim-to-real transfer performance of models trained on our dataset? (3) Can our dataset be used to train different types of models? (4) How can real-world data be leveraged to further improve model performance? We address these questions in Sec.~\ref{ssec:test_different_embodiments},~\ref{ssec:sim_to_real_test},~\ref{ssec:comparison_different_models} and~\ref{ssec:real_data_finetuning}, respectively.

\subsection{Experimental Setup}
\label{ssec:experimental_setup}

In both simulation and real-world experiments, we impose a limit of 1,000 steps; exceeding this limit is counted as a failure. Both simulation and real-world experiments adopt the same frame-by-frame evaluation with an early-stopping mechanism: a trial terminates once the garment, after completing both the unfolding and folding stages, is neatly folded and the robot arm has retracted. In simulation, a garment is considered neatly folded if its semantic keypoints (e.g., shoulders and corners) satisfy predefined distance constraints relative to the target folded configuration; we verify that this rule-based criterion is largely consistent with human judgment. In the real world, a garment is considered neatly folded if it is judged by human experts to reach the tidiness level illustrated in Fig.~\ref{fig:test_tshirt}. In the following figures, we approximate the 68\% confidence interval (CI) using $\text{CI}_{68}=\sqrt{p(1-p)/(k-1)}$, where $p$ is the measured success rate and $k$ is the number of trials, and display it as error bars\footnote{In simulation, we set $k=200$, while in real-world experiments we mainly use $k=15$.}. In both simulation and real-world experiments, the background and garments used for evaluation are unseen during training.

\begin{figure}[t]
\centering
\includegraphics[width=1.0\linewidth]{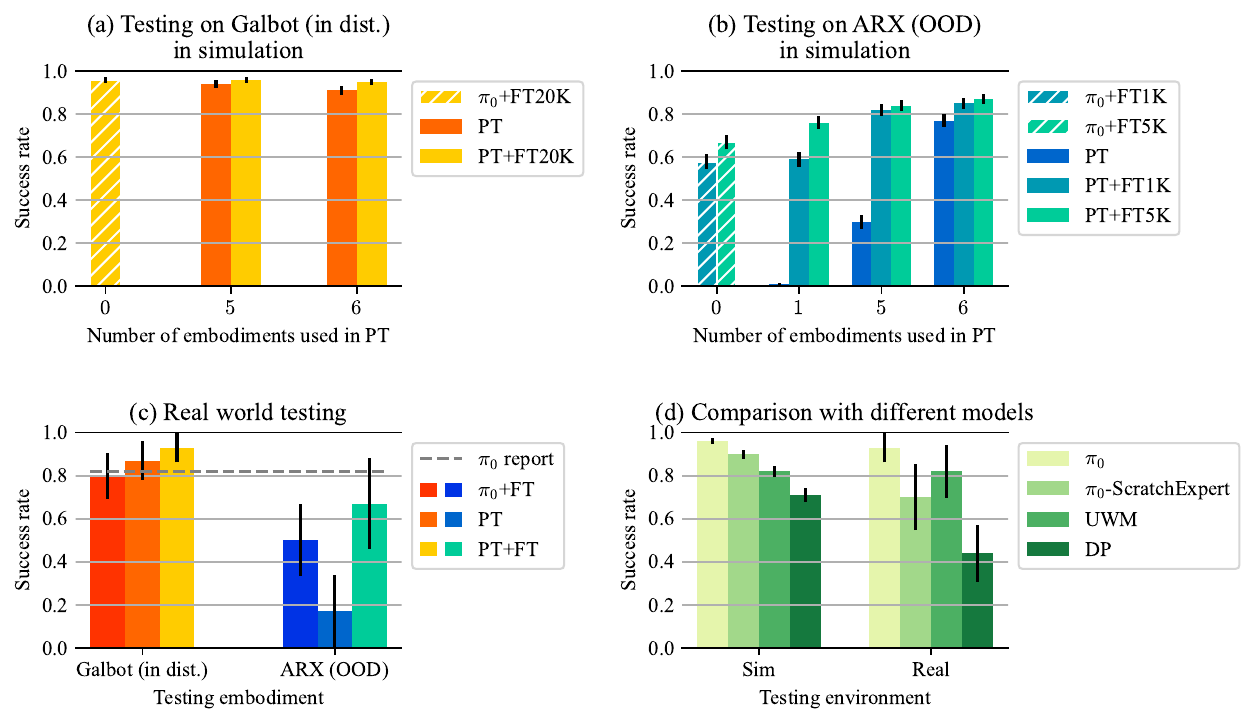}
\caption{
Model evaluation results in simulation and the real world.
}
\label{fig:exp_all_fig}
\vspaceafter
\end{figure}

\subsection{Testing on Different Embodiments}
\label{ssec:test_different_embodiments}

In Fig.~\ref{fig:exp_all_fig}a and Fig.~\ref{fig:exp_all_fig}b, we evaluate model performance across different robot embodiments in simulation. In this experiment, we adopt a pretrained $\pi_0$ model~\cite{black2026pi0visionlanguageactionflowmodel}. Training on FoldNet++ consists of two stages: a cross-embodiment pretraining stage (PT) and an embodiment-specific fine-tuning stage (FT). The x-axis indicates the number of embodiments $n$ used during PT. For the $n=0$ setting, the model is trained for 100K steps from the pretrained $\pi_0$ checkpoint using only the dataset generated for the target embodiment. For all other settings, the model is pretrained on the mixed datasets of the first $n$ embodiments following the ordered sequence: Galbot, AgileX, Franka, UR5e, xArm, and ARX. Each embodiment contributes 20K training episodes during PT, while ``FT~$k$'' in Fig.~\ref{fig:exp_all_fig} denotes fine-tuning on $k$ episodes from the target embodiment. The model is trained for 400K steps during PT, followed by 10K steps during FT.

In Fig.~\ref{fig:exp_all_fig}a, we observe that Galbot, an in-distribution embodiment, consistently achieves high success rates, indicating that our model exhibits strong zero-shot performance on in-distribution embodiments. In addition, fine-tuning in the FT stage provides a modest improvement in success rates. In Fig.~\ref{fig:exp_all_fig}b, for embodiments unseen during the pretraining stage, the model demonstrates a certain degree of cross-embodiment generalization (PT, $n=5$). When $n=6$, ARX is included in the pretraining stage, resulting in a substantial increase in success rate. We also observe that embodiment-specific fine-tuning leads to substantial performance improvements (FT, $n=1,5$). Moreover, as the number of embodiments used during pretraining increases, the post-fine-tuning performance is further improved (FT, $n=0,1,5$); the total amount of pretraining data also increases with $n$, so this improvement reflects the combined effect of data scale and embodiment diversity, which we disentangle in Appendix. After cross-embodiment pretraining, the data requirement for post-training on out-of-distribution embodiments is reduced from 20K to approximately 5K. These results indicate that our model exhibits strong cross-embodiment zero-shot and few-shot capabilities. The lower performance of ARX compared to Galbot is attributed to the smaller workspace of ARX.

\subsection{Sim-to-Real Testing}
\label{ssec:sim_to_real_test}

\begin{wrapfigure}[9]{r}{0.5\textwidth}
\centering
\vspace{-15mm}
\includegraphics[width=1.0\linewidth]{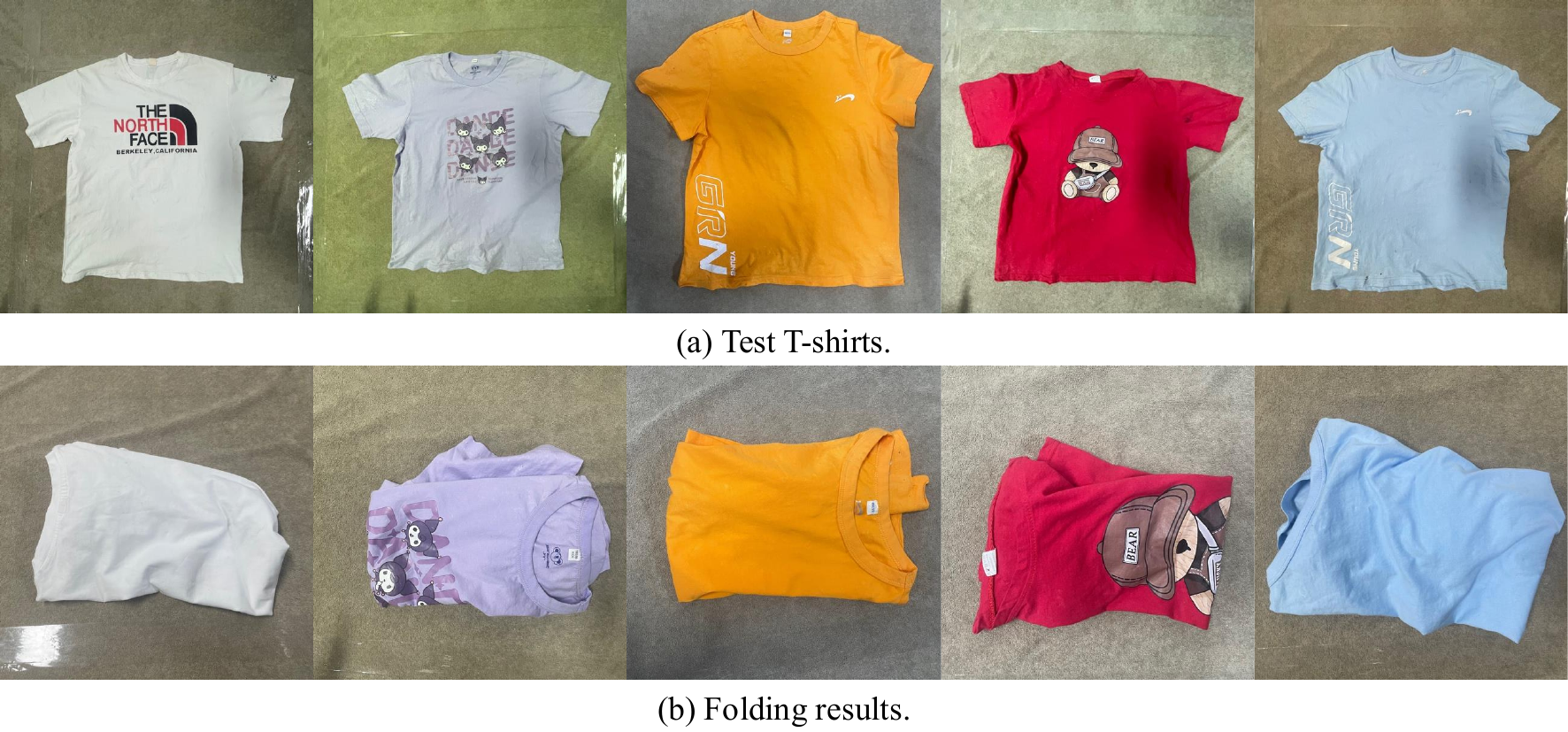}
\caption{
T-shirts used in our real-world experiments and examples of the final folded outcomes.
}
\label{fig:test_tshirt}
\end{wrapfigure}

\begin{figure}[ht]
\centering
\includegraphics[width=1.0\linewidth]{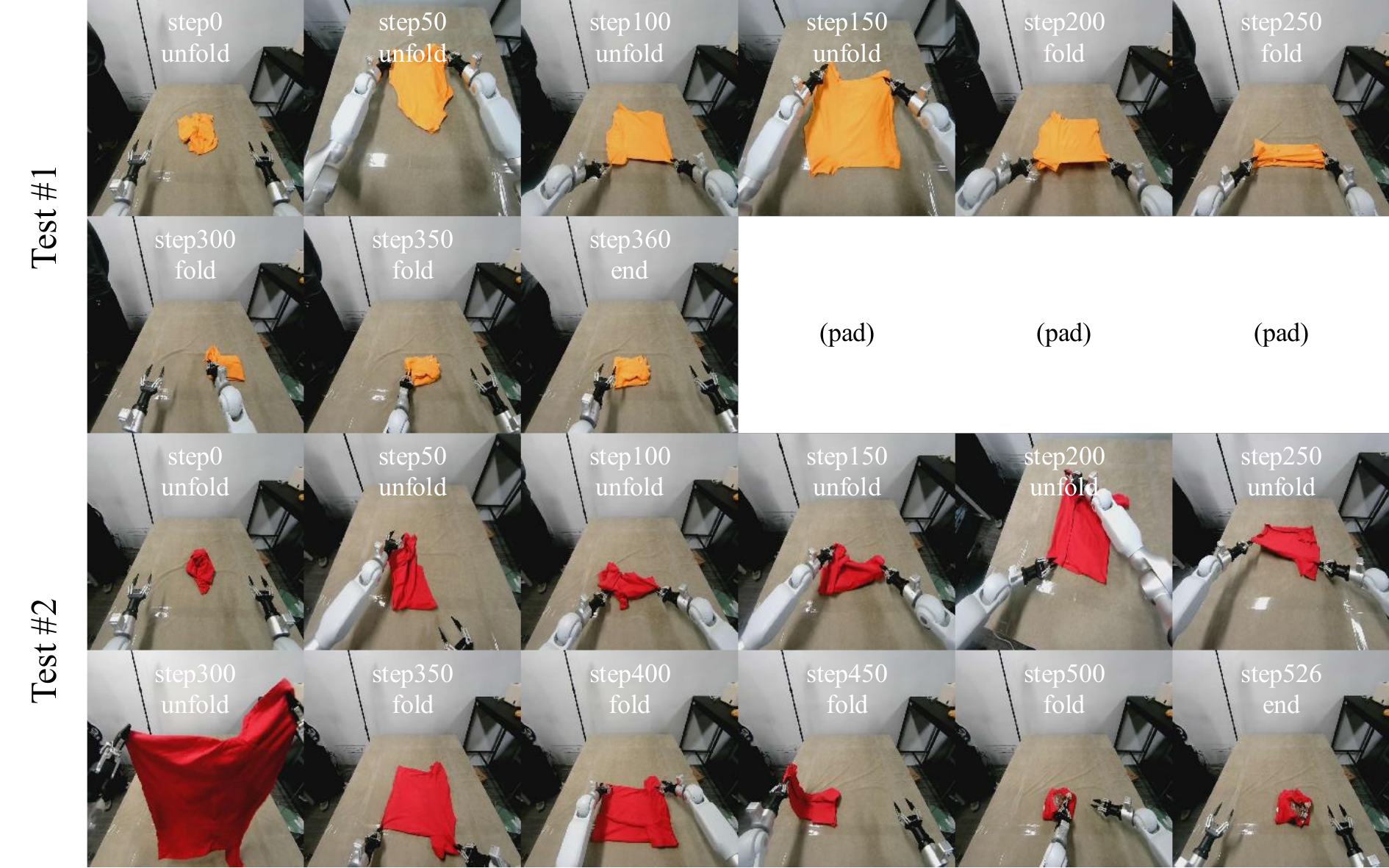}
\caption{Head-view visualization of the model running on Galbot.}
\label{fig:eval_galbot}
\vspaceafter
\end{figure}

\begin{figure}[ht]
\centering
\includegraphics[width=1.0\linewidth]{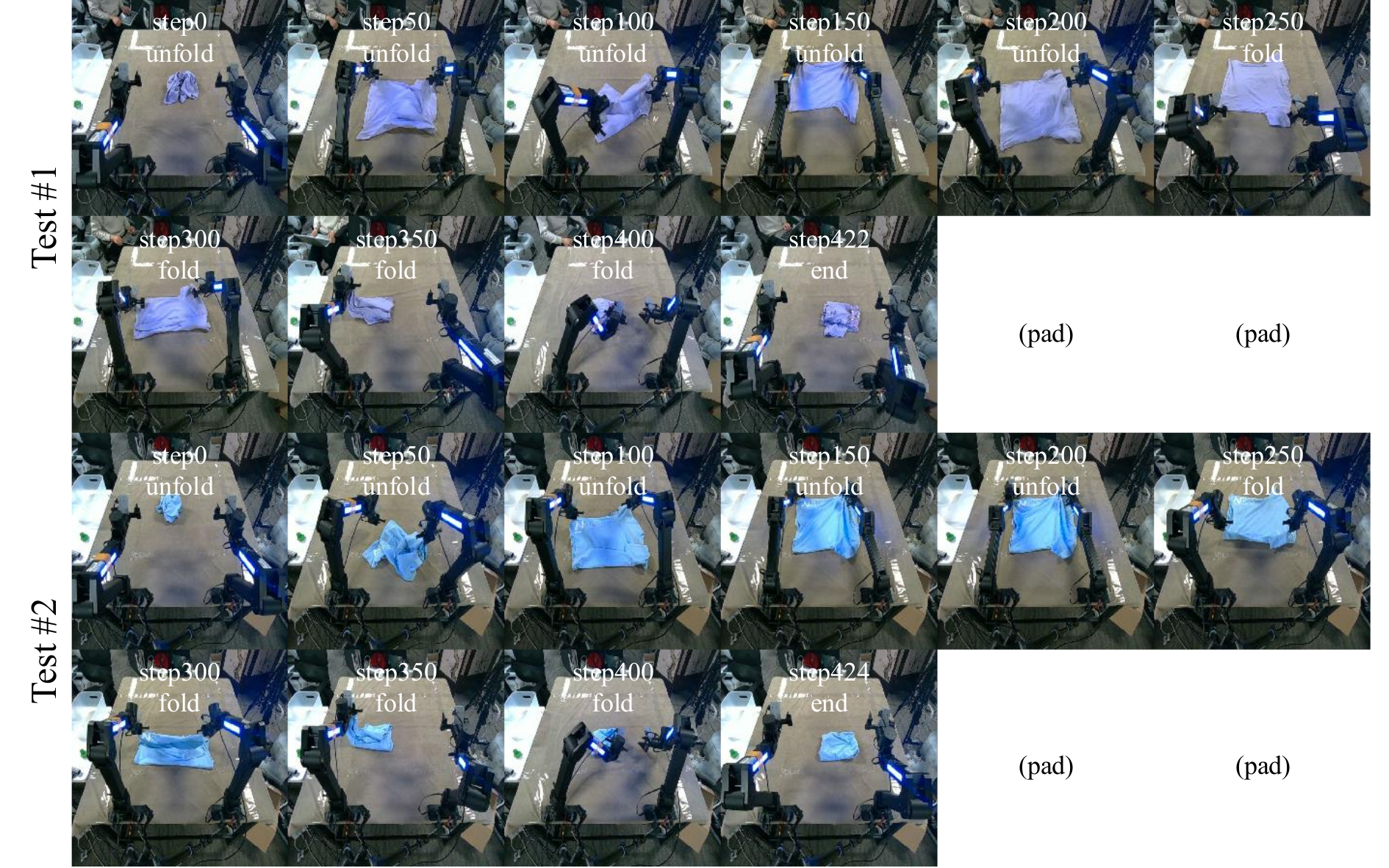}
\caption{Head-view visualization of the model running on ARX.}
\label{fig:eval_arx}
\vspaceafter
\end{figure}

We directly deploy the model to real-world Galbot (an in-distribution embodiment) and ARX (an out-of-distribution embodiment) without any real-world fine-tuning. The experimental results are shown in Fig.~\ref{fig:exp_all_fig}c. The test T-shirts and examples of final folded results are shown in Fig.~\ref{fig:test_tshirt}. Examples of head-view images from real-world executions on Galbot and ARX are shown in Fig.~\ref{fig:eval_galbot} and Fig.~\ref{fig:eval_arx}, respectively.

In Fig.~\ref{fig:exp_all_fig}c, ``$\pi_0$+FT'' denotes training with FT only, ``PT'' denotes training with PT only, and ``PT+FT'' denotes training with both PT and FT. To evaluate performance on an out-of-distribution embodiment, during the PT stage we use the model trained on five embodiments excluding ARX, rather than the full set of six embodiments. For the fine-tuned models, 20K embodiment-specific episodes are used for Galbot and 5K for ARX, respectively.

The results for ``PT'' demonstrate that our cross-embodiment model exhibits zero-shot capability across different embodiments. The results for ``PT+FT'' show that, after cross-embodiment pretraining, additional fine-tuning on a specific embodiment further improves performance.

\subsection{Comparison with Different Models}
\label{ssec:comparison_different_models}

We further evaluate the effectiveness of our dataset across different model architectures. In this experimental section, we use 20K Galbot data for training. We select representative models from three mainstream categories (VLA, WAM, and VA) for comparison. Specifically, we evaluate the pretrained $\pi_0$ model ($\pi_0$)~\cite{black2026pi0visionlanguageactionflowmodel}, a variant with the same architecture as $\pi_0$ but initialized only with VLM weights while the action expert module is trained from scratch ($\pi_0$ ScratchExpert), the unified world model (UWM)~\cite{zhu2025uwm}, and the diffusion policy (DP)~\cite{chi2023diffusionpolicy}. We train the $\pi_0$ model for 100K steps, while UWM and DP are trained for 400K steps.

From Fig.~\ref{fig:exp_all_fig}d, we observe that for ``$\pi_0$ ScratchExpert'', although the model is not pretrained on large-scale multi-task robotic data, it still achieves strong performance using only the FoldNet++ dataset. For smaller models such as UWM and DP, training on our dataset also yields strong success rates in both simulation and real-world settings.

\subsection{Fine-tuning with Real Data}
\label{ssec:real_data_finetuning}

Real-world data can further improve model performance in specific scenarios. We additionally collect 1K teleoperated demonstration trajectories on the test T-shirts using Galbot. We observe that a typical failure case in real-world experiments occurs when the robot repeatedly attempts to grasp the same location on the garment. Accordingly, we adopt two sampling strategies for the teleoperated data during fine-tuning: uniformly sampling from the entire trajectory, denoted as ``FTRD-Uniform'', and sampling only a short segment of actions before grasping, denoted as ``FTRD-Grasp''.

The experimental results are summarized in Tab.~\ref{tab:real_data_finetuning}. We report success rate (SR), average number of execution steps (Steps), and the number of failed grasps (FG) for three different models. 

The results show that ``FTRD-Uniform'' performs worse than the ``Only Sim Data'' baseline. We attribute the inferior performance of ``FTRD-Uniform'' to the discrepancy between the teleoperated action distribution and the rule-based demonstrations. With only a limited amount of real-world data, uniformly sampling from full teleoperated trajectories may introduce inconsistent intermediate manipulation behaviors during fine-tuning.

\begin{wraptable}{r}{0.55\textwidth}
\centering
\begin{tabular}{|c|c|c|c|}
\hline
& SR $\uparrow$ & Steps $\downarrow$ & FG $\downarrow$ \\
\hline
Only Sim Data & 9/10 & 626 & 2.2 \\
FTRD-Uniform & 6/10 & 688 & 2.2 \\
FTRD-Grasp & 9/10 & 458 & 1.8 \\
\hline
\end{tabular}
\caption{Results of fine-tuning with real-world data.}
\label{tab:real_data_finetuning}
\end{wraptable}

In contrast, ``FTRD-Grasp'' restricts real-data supervision to a short pre-grasp segment, where the policy must commit to a precise grasping region. This design avoids introducing diverse intermediate teleoperation behaviors while directly targeting the dominant real-world failure mode: repeated grasp attempts around similar garment regions. By emphasizing states immediately before grasp execution, ``FTRD-Grasp'' improves action stability at the most critical stage of manipulation.


\section{Conclusion}
\label{sec:conclusion}

In this work, we present FoldNet++, a large-scale synthetic dataset for T-shirt folding and unfolding. Using our dataset, models can be directly transferred from simulation to the real world, demonstrating strong generalization capabilities across unseen embodiments, T-shirts, and environments. Extensive experiments in both simulation and real-world settings validate the effectiveness of the proposed dataset.


\section{Limitation}
\label{sec:limitation}

Our current dataset is limited to T-shirts and a single folding style; however, the same rule-based framework can be extended to other garment types and alternative folding procedures in the future. Garment categories with long sleeves introduce additional sleeve-folding manipulations, while zippers and buttons require fine-grained dexterous operations beyond macro-level folding; handling these garment features remains an open problem for future work.


\clearpage
\acknowledgments{We thank the reviewers and area chair for their constructive feedback. We thank Yuxuan Wang and Weiheng Liu for their assistance with the real-world experiments, and Haoran Liu for his help with model training. We also thank Galbot for providing real robot platforms, teleoperation data collection equipment, and computational resources.}


\bibliography{bib/all}  

\clearpage

\appendix

\section{Rule Detail}

\subsection{Unfolding Rules}

\begin{figure}[h]
\centering
\includegraphics[width=1.0\linewidth]{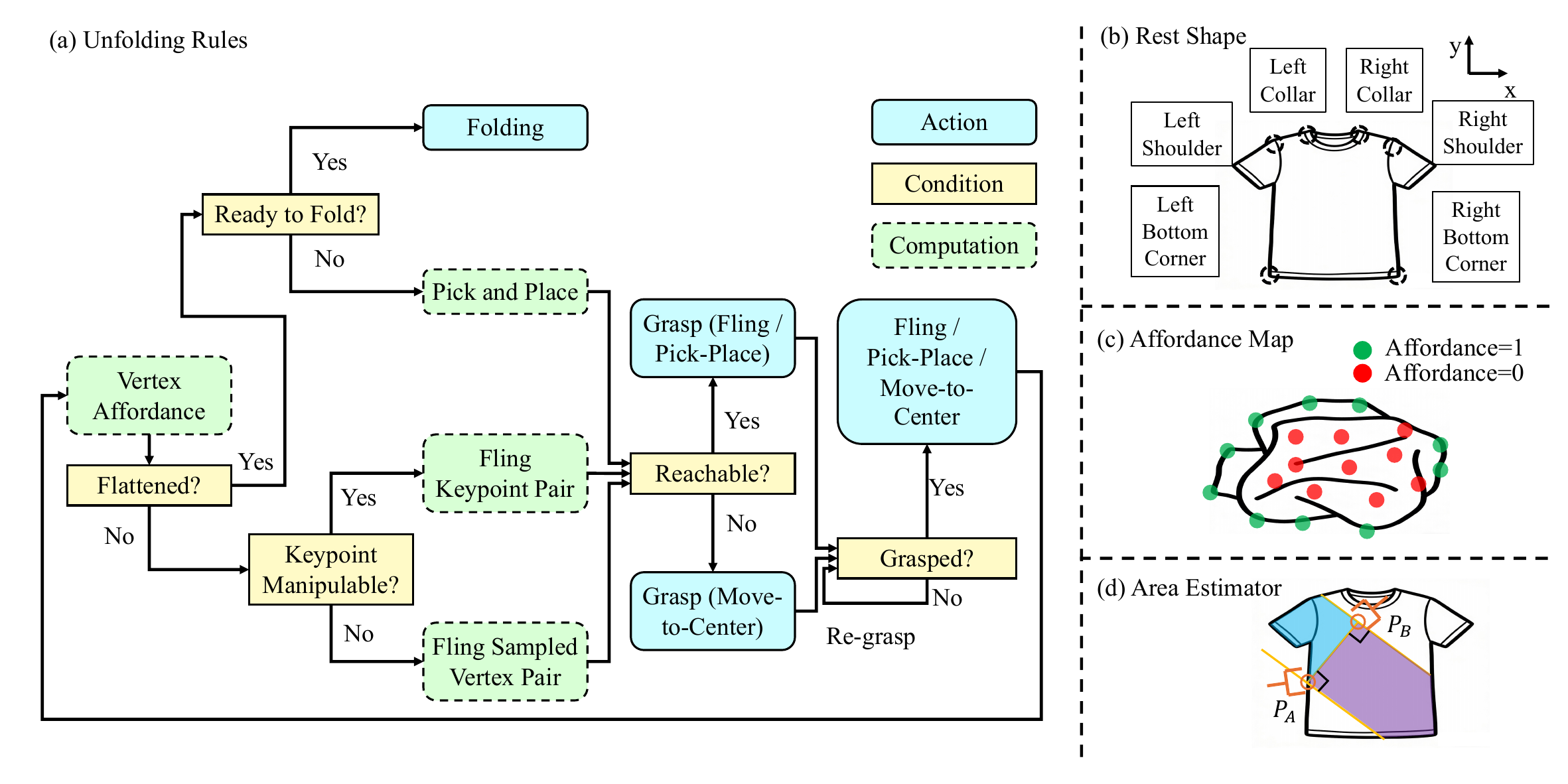}
\caption{
Unfolding rules and relevant concepts. (a) Schematic flowchart of the unfolding rules. To generate unfolding demonstration data, we design state-based rules to flatten the garment from arbitrary initial configurations, preparing it for subsequent folding. (b) The T-shirt in its rest shape, with several main keypoints annotated. (c) A per-vertex affordance map computed by the unfolding rules for a random garment state, indicating whether each vertex is manipulable. (d) Schematic illustration of the area estimation algorithm.
}
\label{fig:unfold_rule_rest_shape}
\end{figure}

The unfolding rules are comparatively more complex and involve multiple conditional branches, as illustrated in Fig.~\ref{fig:unfold_rule_rest_shape}a. In the following description, we abbreviate the main keypoints as follows: LS (left shoulder), RS (right shoulder), LC (left collar), RC (right collar), LB (left bottom corner), and RB (right bottom corner), as shown in Fig.~\ref{fig:unfold_rule_rest_shape}b.

\colorbox{unfoldgreen}{Vertex Affordance.} For each vertex on the garment, we compute its manipulability. In our setting, a value of $1$ indicates that the vertex is manipulable, while a value of $0$ indicates it is not. Since precisely grasping only the top layer of a garment with a gripper is challenging, we assume that when the gripper closes, all vertices within a local neighborhood are grasped simultaneously. A vertex is considered manipulable if and only if the vertices grasped together with it lie within a threshold distance in the $xy$-plane of the rest shape.

The rest shape refers to the undeformed state of the garment, which generally lies in the $xy$-plane with only minor variations along the $z$-axis, as shown in Fig.~\ref{fig:unfold_rule_rest_shape}b. These variations in $z$ arise solely from the front and back layers of the garment.

Formally, let the current position of the $i$-th vertex be denoted as $v(i) \in \mathbb{R}^3$, and its position in the rest shape be denoted as $r(i) \in \mathbb{R}^3$. Let $r_{xy}(i) \in \mathbb{R}^2$ denote the projection of $r(i)$ onto the $xy$-plane. We define
\[
G = \left\{ j \;\middle|\; \|v(j) - v(i)\| < \SI{0.02}{\meter} \right\}
\]
as the set of vertices grasped together with the $i$-th vertex. The affordance label of vertex $i$ is defined as $1$ if and only if
\[
\max_{j \in G} \|r_{xy}(i) - r_{xy}(j)\| < \SI{0.08}{\meter}.
\]

An example of an affordance map is shown in Fig.~\ref{fig:unfold_rule_rest_shape}c. In general, for a random garment state, boundary regions are often manipulable, whereas vertices in stacked regions are not. For example, when the left sleeve is stacked on top of the right sleeve, grasping a vertex on the sleeve results in both sleeves being picked up simultaneously. Since the left and right sleeves are far apart in the rest shape, neither sleeve is considered manipulable under our criterion. In practice, this is consistent with observation: grasping both sleeves simultaneously makes it difficult for a subsequent fling action to properly flatten the garment.

The affordance map computation method described here serves only as one possible implementation, and alternative approaches can also be used to compute the affordance map. In simulation, we find that our computation method effectively guides the subsequent fling and pick-and-place operations.

\colorbox{unfoldyellow}{Flattened?} First, we determine whether the garment has already been flattened. The criteria are that all keypoints LS, RS, LC, RC, LB, and RB are manipulable, and that their average deviation from the rest shape (up to a rigid-body transformation) does not exceed \SI{0.03}{\meter}. In addition, the current garment orientation must differ from the required initial orientation for folding by less than $22.5^\circ$.

\colorbox{unfoldyellow}{Ready to Fold?} When the garment is already flattened and its orientation error is small, we further check whether it is properly centered. If the maximum positional deviation between the current locations of the LC, RC, LB, and RB keypoints and their corresponding target positions (when the garment is centered at the origin of the world coordinate system) is less than \SI{0.08}{\meter}, we switch to the folding strategy. Otherwise, a pick-and-place action is used to reposition the garment.

\colorbox{unfoldblue}{Folding.} We consider four ($2\times2$) possible initial configurations for folding a T-shirt: the neckline facing left or right, and the front side facing up or down. Once the initial configuration is determined, we adjust the sign conventions of the corresponding parameters in the rules described in Sec.~\ref{ssec:folding_rules} and proceed with the folding process.

\colorbox{unfoldgreen}{Pick and Place.} For the four keypoints LB, RB, LS, and RS, if the garment needs to be dragged forward, we select the two keypoints closer to the front; if it needs to be dragged backward, we select the two keypoints closer to the rear.

\colorbox{unfoldyellow}{Keypoint Manipulable?} We predefine several keypoint pairs. According to priority, the highest-priority vertex pairs are (LB, LS) and (RB, RS). Flinging any of these pairs can simultaneously flatten the garment and adjust its orientation for subsequent folding. In our setting, both front-side-up and back-side-up configurations are allowed. The next-priority pairs are (LB, RB) and (LS, RS). These pairs can also achieve flattening, while orientation adjustments are deferred to the next fling action.

\colorbox{unfoldgreen}{Fling Keypoint Pair.} We select keypoint pairs to fling according to the predefined priority.

\colorbox{unfoldgreen}{Fling Sampled Vertex Pair.} If none of the predefined keypoint pairs are manipulable, we sample candidate pairs from all manipulable vertex pairs. For each pair $(P_A, P_B)$, we estimate the area after a single fling operation when grasping these two vertices, as illustrated in Fig.~\ref{fig:unfold_rule_rest_shape}d. To estimate this area, we first map all vertices back to the rest shape. Then, we construct perpendicular lines to the segment $P_A P_B$ passing through $P_A$ and $P_B$, respectively. Finally, we compute the enclosed areas and take the larger one as the estimated post-fling area, i.e., $A = \max(A_{\text{blue}}, A_{\text{purple}})$. This value is used as a score, and the pair with the highest score among the samples is selected. In addition, we apply a penalty if the distance from the robot arm to the candidate pair exceeds a predefined threshold based on the robot-specific workspace. We also enforce a minimum distance of \SI{0.1}{\meter} between the two vertices to prevent gripper collisions during grasping.

\colorbox{unfoldyellow}{Reachable?} For both fling and pick-and-place operations, we grasp two points simultaneously using both hands. We then check whether both points lie within the robot’s workspace. If the grasping pose lies outside the workspace, we first reposition the garment toward the center of the workspace.

\colorbox{unfoldblue}{Grasp (Fling / Pick-Place).} Move the robot’s grippers to the target grasping positions and execute the grasp. The grasping positions are determined based on the previously planned fling or pick-and-place actions, as shown in steps 2, 4, 9, 14, and 17 in Fig.~\ref{fig:fold_vis_unfold_vis}b.

\colorbox{unfoldblue}{Grasp (Move-To-Center).} Move the robot’s grippers to the target grasping positions and execute the grasp. The grasping positions are located on the left and right sides of the garment, and the target positions are approximately centered within the robot’s workspace.

\colorbox{unfoldyellow}{Grasped?} We check whether the two target points have been successfully grasped. If the grasp succeeds, we proceed to the subsequent steps; otherwise, we retry the grasp. This re-grasping process also follows the KG-DAgger strategy from prior work~\cite{chen2025foldnetlearninggeneralizableclosedloop}, as shown in steps 3 and 16 in Fig.~\ref{fig:fold_vis_unfold_vis}b.

\colorbox{unfoldblue}{Fling / Pick-Place / Move-To-Center.} For the \textit{Fling} action, we first lift the garment to a fixed height, as shown in steps 5 and 10 in Fig.~\ref{fig:fold_vis_unfold_vis}b. We then stretch it laterally, as shown in steps 6 and 11 in Fig.~\ref{fig:fold_vis_unfold_vis}b. The stretch distance is computed based on the two grasp points in the rest shape, preventing excessive stretching that could damage the garment. This state-based formulation allows the information to be distilled into a vision-based model, rather than relying on fixed hyperparameters used in the action primitives of prior work~\cite{canberk2022clothfunnelscanonicalizedalignmentmultipurpose,xue2023unifolding}. After stretching, we execute the fling along an S-shaped trajectory from $(y_1,z_1)$ to $(y_2,z_2)$:
\[
z = z_2 + (z_1 - z_2)\frac{\cos\left(\frac{y - y_1}{y_2 - y_1}\pi\right) + 1}{2},
\]
as shown in steps 7 and 12 in Fig.~\ref{fig:fold_vis_unfold_vis}b. For \textit{Pick-Place} and \textit{Move-to-Center} actions, we directly move both hands from the initial positions to the target positions, as shown in steps 15 and 18 in Fig.~\ref{fig:fold_vis_unfold_vis}b.

We compare several trajectory forms, including straight lines and elliptical arcs, and find that the cosine-shaped trajectory performs better. Even under large model inference latency (about \SI{0.4}{\second}), using this trajectory for the fling operation can still flatten the garment effectively.

\subsection{KG-DAgger}

We simplify the KG-DAgger procedure proposed in~\cite{chen2025foldnetlearninggeneralizableclosedloop} by adopting an offline variant. Specifically, during data generation, we intentionally inject noise to simulate policy prediction errors, enabling the model to acquire error-recovery capabilities from the initial training stage. As a result, our pipeline can directly generate regrasping demonstrations, as illustrated in step 5 of Fig.~\ref{fig:fold_vis_unfold_vis}a and steps 3 and 16 of Fig.~\ref{fig:fold_vis_unfold_vis}b.

In step 5 of Fig.~\ref{fig:fold_vis_unfold_vis}a, the left gripper fails to grasp the garment while the right gripper succeeds. The policy therefore learns to detect the failed grasp and regrasp the garment with the left gripper. Similarly, in steps 3 and 16 of Fig.~\ref{fig:fold_vis_unfold_vis}b, the policy learns to recover by regrasping the garment with the right gripper.

Consistent with the findings in~\cite{chen2025foldnetlearninggeneralizableclosedloop}, introducing KG-DAgger enables the model to reattempt grasping after grasp failures, substantially improving model robustness.

\subsection{Trajectory Generation}
\label{ssec:trajectory_generation}

Applying the rules described in Sec.~\ref{ssec:folding_rules} and~\ref{ssec:unfolding_rules} yields only the initial and target end-effector poses of the robot. The trajectory for fling actions follows a cosine profile, while for all other actions we use uniformly spaced linear interpolation in both translation and rotation between the two poses, enabling the end effector to move along a straight path from the initial to the target pose. In Euclidean space, the step length for fling actions is set to \SI{0.09}{\meter}, while the step size for all other actions is set to \SI{0.03}{\meter}. Subsequently, inverse kinematics (IK) is used to solve for the robot’s joint angles, which are then executed in the simulator.

If the robot has additional leg degrees of freedom, we heuristically compute the leg joint angles during the grasping phase of the unfolding stage, where workspace constraints are most stringent. Specifically, when the grasp point has a large $y$-coordinate, the robot leans forward; when the grasp point has a large $x$-coordinate, the robot rotates to the right. If the robot does not have leg degrees of freedom, or if it is not in the grasping phase of unfolding, all non-arm joints are set to their default initial configurations. During IK computation, we simplify the system as two independent manipulators and solve only for the arm degrees of freedom, without considering the leg joints.

\section{Simulation Detail}

\subsection{Physical Parameters}

The table below lists the main physical parameters. $\mathcal{U}(a,b)$ denotes uniform sampling within the interval $[a,b]$.

\begin{longtable}{|>{\centering\arraybackslash}p{2.0cm}|>{\centering\arraybackslash}p{3.5cm}|>{\centering\arraybackslash}p{7.0cm}|}
\hline
Parameter & Value & Description \\ \hline
\endfirsthead

\hline
\endhead
dt\_step & \SI{0.2}{\second} & Time interval between consecutive actions output by the policy. \\ \hline
dt\_robot & \SI{8}{\milli\second} & Simulation time step for a single robot update. \\ \hline
dt\_cloth & \SI{0.8}{\milli\second} & Simulation time step for a single cloth update. \\ \hline
thickness & \SI{0.8}{\milli\meter} & T-shirt thickness, representing the self-collision thickness of the cloth. \\ \hline
density & \SI{0.2}{\kilo\gram\cdot\meter^{-2}} & Cloth area density. \\ \hline
scale & \SI{0.57}{\meter}$\times\mathcal{U}(0.9,1.1)$ & Geometric mean of the T-shirt length and width. \\ \hline
stretching stiffness & \SI{70}{\kilo\gram\cdot\second^{-2}}$\times\mathcal{U}(0.8,1.2)$ & Cloth stiffness against stretching deformation. \\ \hline
bending stiffness & \SI{7e-7}{\kilo\gram\cdot\meter^{2}\cdot\second^{-2}} $\times\mathcal{U}(0.8,1.2)$ & Cloth stiffness against bending deformation. \\ \hline
table friction coefficient & $0.5\times\mathcal{U}(0.8,1.2)$ & Friction coefficient between the table and the cloth.  \\ \hline
cloth friction coefficient & $1.0\times\mathcal{U}(0.8,1.2)$ & Self-friction coefficient of the cloth. \\ \hline
picker size & $[-0.01, +0.01]\times[-0.008,+0.008]\times[-0.013,+0.030]$\SI{}{\meter} & Cloth vertices within this box in the EEF frame are attached to the EEF when the gripper closes. \\ \hline

\end{longtable}

\subsection{Camera Parameters}

Except for the head camera of Galbot, all other head and wrist cameras use the Intel RealSense D405. The head camera resolution is $320 \times 240$. For Galbot’s head camera, $f_x$ and $f_y$ are 144, while for the other D405 head cameras, $f_x$ and $f_y$ are 195. All wrist-mounted D405 cameras have a resolution of $320 \times 180$ with $f_x = f_y = 163$. 

In simulation, the camera intrinsics $(f_x, f_y, c_x, c_y)$ are perturbed by random noise sampled from a uniform distribution $\mathcal{U}(-16, +16)$. The camera extrinsics are randomized within a translation range of \SI{0.02}{\meter} and a rotation range of $0.1$ radians.

\subsection{Other Parameters}

\textbf{Table.} The table is approximately aligned with the $z=0$ plane, while its pose is randomized within a translation range of \SI{0.03}{\meter} and a rotation range of $0.03$ radians.

\textbf{Initial pose.} Before the policy starts, the left and right EEF poses are initialized to $xyzrpy=[\mp0.35, -0.2, 0.15, \pi/2, 0, 0]$, respectively. The position is measured in meters, and the orientation is represented using RPY Euler angles. This initial pose is randomized within a translation range of \SI{0.05}{\meter} and a rotation range of $0.1$ radians.

\textbf{Robot pose.} The base poses of all robot platforms are listed in the table below. For integrated robot platforms, the base pose is represented by a single value. For platforms composed of two robotic arms, the base pose specifies the poses of the left and right arms, respectively. During evaluation, an additional perturbation within a translation range of \SI{0.02}{\meter} and a rotation range of $0.02$ radians is applied to the base poses of dual-arm platforms.

\begin{longtable}{|>{\centering\arraybackslash}p{2.0cm}|>{\centering\arraybackslash}p{6.0cm}|}
\hline
Robot name & Robot base \\ \hline
\endfirsthead

\hline
\endhead
Galbot & $[0, -0.8, -0.72], [0, 0, \pi/2]$ \\ \hline
AgileX & $[0, -0.8, -0.75], [0, 0, \pi/2]$ \\ \hline
Dual Franka & $[\mp0.30, -0.65, 0], [0, 0, \pi/2]$ \\ \hline
Dual UR5e   & $[\mp0.40, -0.65, 0], [0, 0, \mp\pi/2]$ \\ \hline
Dual xArm   & $[\mp0.35, -0.60, 0], [0, 0, \pi/2]$ \\ \hline
Dual ARX    & $[\mp0.30, -0.55, 0], [0, 0, \pi/2]$ \\ \hline

\end{longtable}

\subsection{Demonstration Generation Cost}

For each robotic embodiment, we generate 20K episodes, resulting in a total of 120K episodes across all embodiments. The training data generation process takes 7 days on a cluster of 96 NVIDIA 4090 GPUs. 

\section{Training Detail}

\subsection{Shared Training Setup}

The action space has a dimensionality of $9 + 9 + 2 + 4 = 24$. Here, $9$ corresponds to the end-effector pose of a single arm, with rotation represented in a 6D continuous form~\cite{Zhou_2019_CVPR}; $2$ corresponds to the gripper actions of the two arms; and $4$ corresponds to the leg joint actions. For robots without leg degrees of freedom, these dimensions are zero-padded. The shared parameters are summarized in the table below.

\begin{longtable}{|>{\centering\arraybackslash}p{4.0cm}|>{\centering\arraybackslash}p{6.0cm}|}
\hline
Parameter & Value \\ \hline
\endfirsthead

\hline
\endhead
Action Horizon & 16 \\ \hline
Proprio Length & 2 \\ \hline
Rollout Length & 2 \\ \hline
Action Representation & relative~\cite{chi2024universal}\\ \hline
Precision & float32 \\ \hline
LR Schedule & cosine with warmup \\ \hline
Warmup Ratio & 0.01 \\ \hline
Freeze Modules & None \\ \hline
Color Jitter & $\text{RandomCrop}(0.95H, 0.95W)$, \text{Resize}$(224, 224)$, $\text{RandomRotation}(-5,+5)$, $\text{ColorJitter}(0.3, 0.4, 0.5)$ \\ \hline

\end{longtable}

Flow Matching Batch Size $N$ is defined as follows: for each target action $A \in \mathbb{R}^{T_a \times D_a}$, we perform only a single forward pass outside the action expert. Within the action expert, the data is replicated $N$ times, and $N$ independent timestep and noise samples are drawn when computing the flow-matching loss, thereby increasing the diversity of timestep and noise samples.

\subsection{$\pi_0$ Model}

The training hyperparameters of the $\pi_0$ model~\cite{black2026pi0visionlanguageactionflowmodel} are listed in the table below. All other settings are kept the same as in the original model.

\begin{longtable}{|>{\centering\arraybackslash}p{4.0cm}|>{\centering\arraybackslash}p{6.0cm}|}
\hline
Parameter & Value \\ \hline
\endfirsthead

\hline
\endhead

Batchsize & 64 \\ \hline
Flow Matching Batchsize & 4 \\ \hline
Learning Rate & 2.5e-5 \\ \hline
Language Instruction & Flatten the T-shirt and then fold it. \\ \hline

\end{longtable}

\subsection{Unified World Model}

The training hyperparameters of the UWM model~\cite{zhu2025uwm} are listed in the table below. All other settings are kept the same as in the original model.

\begin{longtable}{|>{\centering\arraybackslash}p{4.0cm}|>{\centering\arraybackslash}p{6.0cm}|}
\hline
Parameter & Value \\ \hline
\endfirsthead

\hline
\endhead

Obs Length & 1 \\ \hline
Action Loss Weight & 1.0 \\ \hline
Dynamic Loss Weight & 0.0 \\ \hline
Model Size & 346M \\ \hline
Batchsize & 128 \\ \hline
Flow Matching Batchsize & 16 \\ \hline
Learning Rate & 1e-4 \\ \hline

\end{longtable}

\subsection{Diffusion Policy}

The training hyperparameters of the DP model~\cite{chi2023diffusionpolicy} are listed in the table below. All other settings are kept the same as in the original model.

\begin{longtable}{|>{\centering\arraybackslash}p{4.0cm}|>{\centering\arraybackslash}p{6.0cm}|}
\hline
Parameter & Value \\ \hline
\endfirsthead

\hline
\endhead

Obs Encoder & Resnet50 \\ \hline
UNet Dimensions & $[512, 1024, 2048]$ \\ \hline
Model Size & 269M \\ \hline
Batchsize & 128 \\ \hline
Flow Matching Batchsize & 16 \\ \hline
Learning Rate & 1e-4 \\ \hline

\end{longtable}

\subsection{Real data Finetuning}

The ``Only Sim Data'' model is pretrained using six robotic embodiments and subsequently fine-tuned on Galbot, with training data consisting exclusively of synthetic data. For ``FTRD-Uniform'' and ``FTRD-Grasp'', we fine-tune the model using a $1{:}1$ sampling ratio between synthetic and real-world data. Starting from the ``Only Sim Data'' model, ``FTRD-Uniform'' is trained for 2K steps with a learning rate of 2.5e-6, while ``FTRD-Grasp'' is trained for 20K steps with a learning rate of 2.5e-5. These training durations are selected based on empirical evaluations across multiple checkpoints and correspond to the best-performing configurations.

\section{Control Detail}

\begin{figure}[h]
    \centering
    \includegraphics[width=1.0\linewidth]{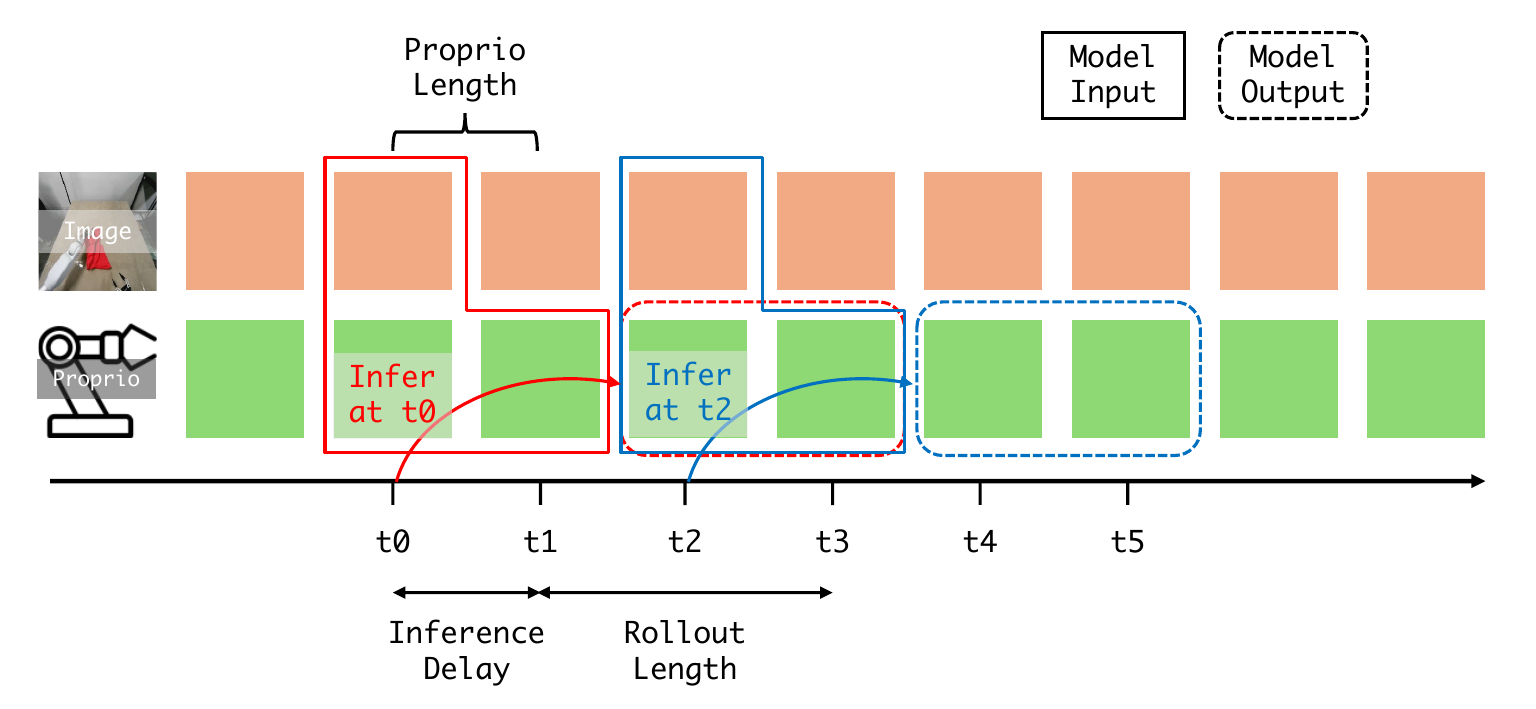}
    \caption{Real-time Chunking.}
    \label{fig:rtc}
\end{figure}

For all models, we adopt Training-Time RTC control~\cite{black2025trainingtimeactionconditioningefficient} to enable asynchronous inference, as shown in Fig.~\ref{fig:rtc}.

We solve IK using an optimization-based approach built upon Newton iterations and the end-effector Jacobian. After the model predicts the target end-effector poses, we convert them to absolute poses and update the leg joints if leg actions are available. We then initialize the IK optimization with $q_i = \lambda q_p + (1 - \lambda) q_0$, where $\lambda = 0.95$ is a smoothing coefficient, $q_p$ denotes the previous joint configuration, and $q_0$ is a manually specified reference pose in which both grippers lie within a dexterous workspace. This initialization strategy gradually biases the solution toward $q_0$ during each IK step. While this approach provides limited benefits for 6-DoF robotic arms such as ARX, it significantly improves robustness for redundant 7-DoF manipulators such as Galbot.

For gripper actions, we define $0$ as open and $1$ as closed. All gripper actions are discretized into three bins: $0$, $0.5$, and $1.0$.

\section{Test Garment Details}

For the real T-shirts used in evaluation, the length ranges from \SI{49}{\centi\meter} to \SI{62}{\centi\meter}, and the width ranges from \SI{60}{\centi\meter} to \SI{73}{\centi\meter}. Across all tested T-shirts, we do not observe any garment-specific failure pattern. The synthetic 1K T-shirts follow the same short-sleeve garment distribution as in FoldNet~\cite{chen2025foldnetlearninggeneralizableclosedloop}, with variations in both geometry and visual texture.

\section{Failure Mode}

\begin{figure}[h]
\centering

\begin{subfigure}{0.48\textwidth}
\centering
\includegraphics[width=\textwidth]{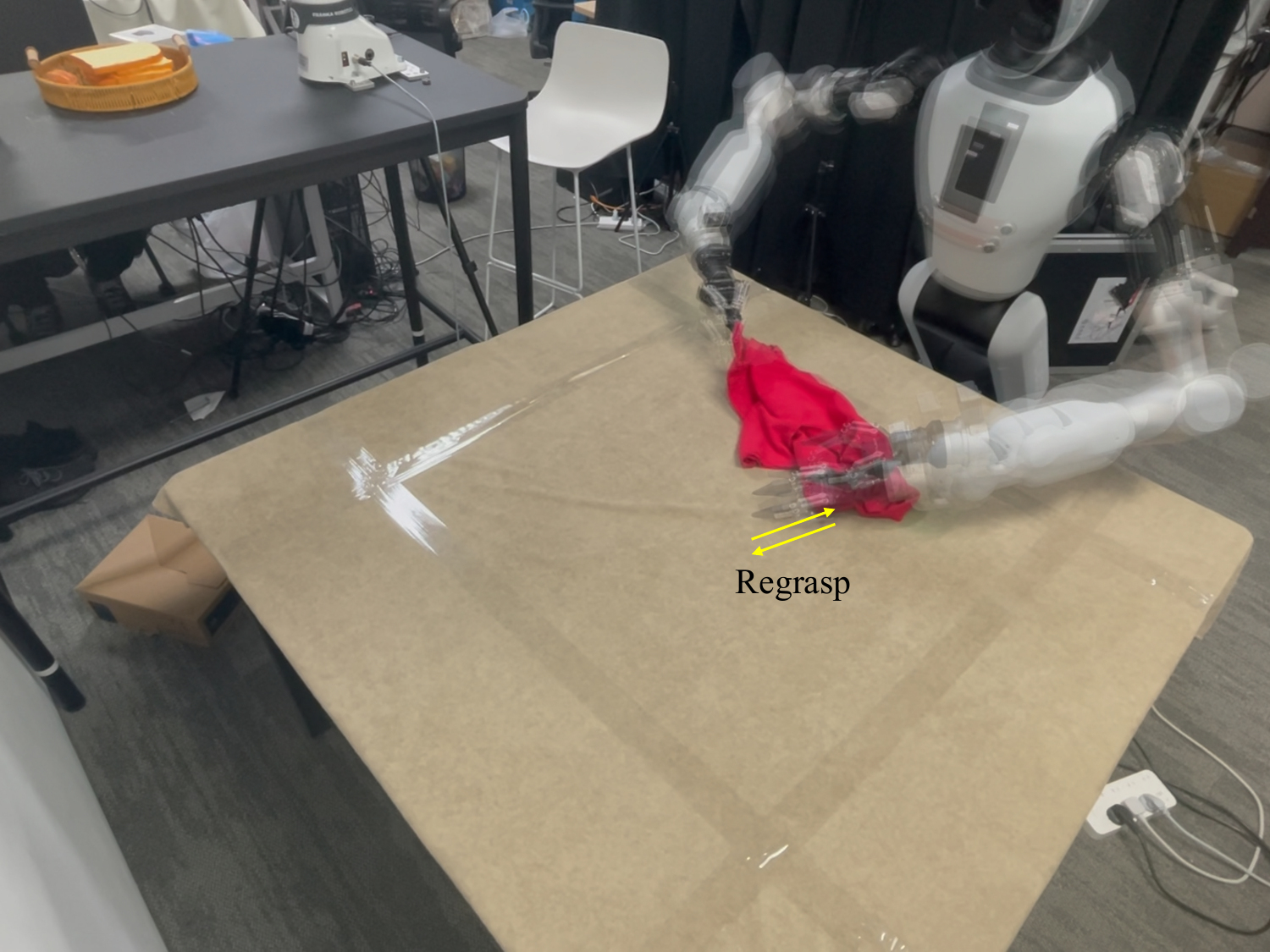}
\caption{The robot repeatedly attempts to grasp but fails, resulting in excessive time consumption.}
\end{subfigure}
\hfill
\begin{subfigure}{0.48\textwidth}
\centering
\includegraphics[width=\textwidth]{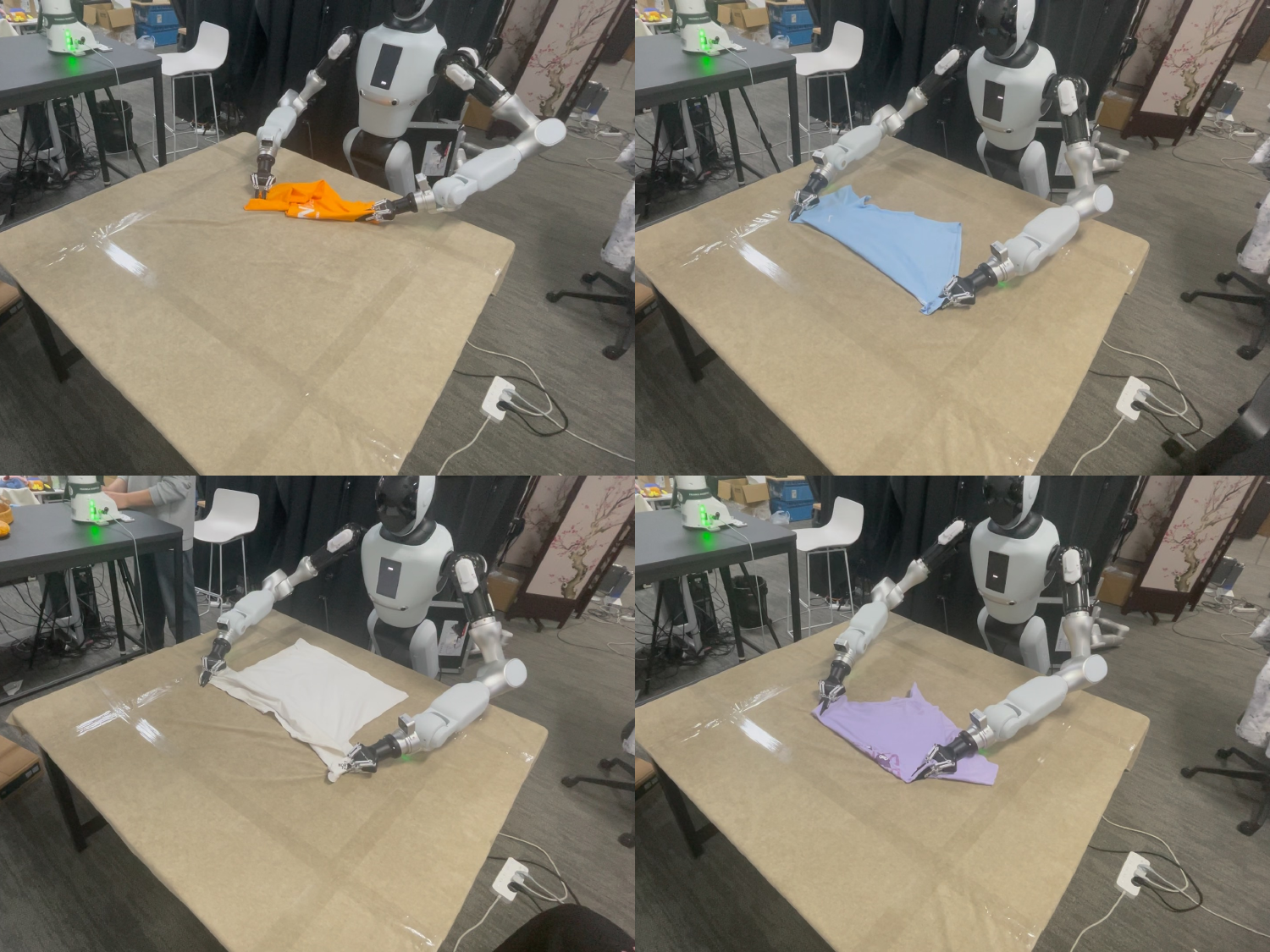}
\caption{The selected grasping point for the fling motion is suboptimal.}
\end{subfigure}

\caption{Typical failure modes.}
\label{fig:failure_mode}
\end{figure}

Cases classified as ``unsuccessful'' mainly include unsafe robot behaviors (e.g., self-collisions) and exceeding the maximum step limit. The failure modes discussed here primarily refer to suboptimal intermediate behaviors during execution, rather than final task failure. Several typical real-world failure cases are illustrated in Fig.~\ref{fig:failure_mode}.

Since our model operates in a closed-loop manner, it can detect such failures and automatically attempt recovery behaviors. Most failures are temporary, and given sufficient time, the robot can usually restore the garment to a normal state. In the future, collecting more diverse training data may further improve efficiency and reduce the frequency of such failures.

\section{Additional Experiments}
\label{sec:additional_experiments}

\paragraph{Fixed-data-budget ablation.} As noted in Sec.~\ref{ssec:test_different_embodiments}, the improvement in post-fine-tuning performance with increasing $n$ reflects the combined effect of a larger total amount of pretraining data and greater embodiment diversity. To disentangle these two factors, we conduct an additional ablation in which the total number of pretraining episodes is fixed at 20K, rather than scaling with $n$, followed by fine-tuning with 1K episodes. As shown in Tab.~\ref{tab:fixed_budget_ablation}, as $n$ increases from 1 to 3 to 5, the success rate improves from 59\% to 64\% to 65\%, respectively, indicating that embodiment diversity provides an additional benefit even when the overall scale of pretraining data remains the dominant factor.

\begin{table}[h]
\centering
\begin{tabular}{|c|c|c|c|}
\hline
$n$ & 1 & 3 & 5 \\
\hline
SR $\uparrow$ & 59\% & 64\% & 65\% \\
\hline
\end{tabular}
\caption{Success rate as a function of the number of pretraining embodiments $n$, under a fixed pretraining budget of 20K episodes followed by fine-tuning on 1K episodes.}
\label{tab:fixed_budget_ablation}
\end{table}

\paragraph{Real-only baseline and fine-tuning hyperparameters.} To further contextualize the results in Sec.~\ref{ssec:real_data_finetuning}, we additionally train a real-only baseline using exclusively the 1K teleoperated real-world demonstrations, without any synthetic pretraining. We also examine the sensitivity of ``FTRD-Uniform'' and ``FTRD-Grasp'' to the fine-tuning configuration. Under a shared, relatively aggressive configuration, ``FTRD-Uniform'' and ``FTRD-Grasp'' achieve success rates of 4/10 and 9/10, respectively, as shown in Tab.~\ref{tab:realonly_ftrd}. The lower performance of ``FTRD-Uniform'' under this configuration arises from the distribution mismatch between simulation and real-world data: applying an aggressive fine-tuning configuration to a limited amount of real-world data leads to overfitting and degrades the learned policy. Adopting the more conservative fine-tuning configuration reported in Tab.~\ref{tab:real_data_finetuning} improves ``FTRD-Uniform'' from 4/10 to 6/10.

\begin{table}[h]
\centering
\begin{tabular}{|c|c|}
\hline
Method & SR $\uparrow$ \\
\hline
Real-only & 2/10 \\
FTRD-Uniform (aggressive) & 4/10 \\
FTRD-Grasp (aggressive) & 9/10 \\
\hline
\end{tabular}
\caption{Real-only baseline and fine-tuning results under an aggressive fine-tuning configuration.}
\label{tab:realonly_ftrd}
\end{table}

\paragraph{Wrist camera and physics engine ablation.} A direct success-rate comparison with FoldNet is not meaningful, as the two methods target different tasks. Instead, we ablate two key design choices within the FoldNet++ pipeline in simulation. Removing wrist-mounted cameras reduces the success rate from 96\% to 58\% (Tab.~\ref{tab:wrist_camera_ablation}), indicating that wrist cameras substantially improve grasping accuracy. Style3D is likewise a necessary component of our pipeline for generating valid demonstrations; however, it is difficult to ablate directly, as PyFlex is unable to reliably complete the more complex unfolding-to-folding pipeline required by our task.

\begin{table}[h]
\centering
\begin{tabular}{|c|c|}
\hline
Camera configuration & SR $\uparrow$ \\
\hline
With wrist camera & 96\% \\
Without wrist camera & 58\% \\
\hline
\end{tabular}
\caption{Effect of removing wrist-mounted cameras on success rate.}
\label{tab:wrist_camera_ablation}
\end{table}

\paragraph{Galbot as OOD embodiment.} Using a fixed pretraining budget of 20K episodes across $n=5$ embodiments followed by fine-tuning on 1K target-embodiment episodes, we compare Galbot and ARX in simulation when each is held out of pretraining as the out-of-distribution embodiment. Galbot achieves a success rate of 87\%, compared with 63\% for ARX (Tab.~\ref{tab:galbot_arx_ood}), suggesting that ARX's smaller workspace is a primary factor behind its lower performance.

\begin{table}[h]
\centering
\begin{tabular}{|c|c|}
\hline
Embodiment & SR $\uparrow$ \\
\hline
Galbot as OOD & 87\% \\
ARX as OOD & 63\% \\
\hline
\end{tabular}
\caption{Success rate comparison between Galbot and ARX under a fixed pretraining and fine-tuning budget.}
\label{tab:galbot_arx_ood}
\end{table}

\begin{figure}[h]
\centering
\includegraphics[width=\textwidth]{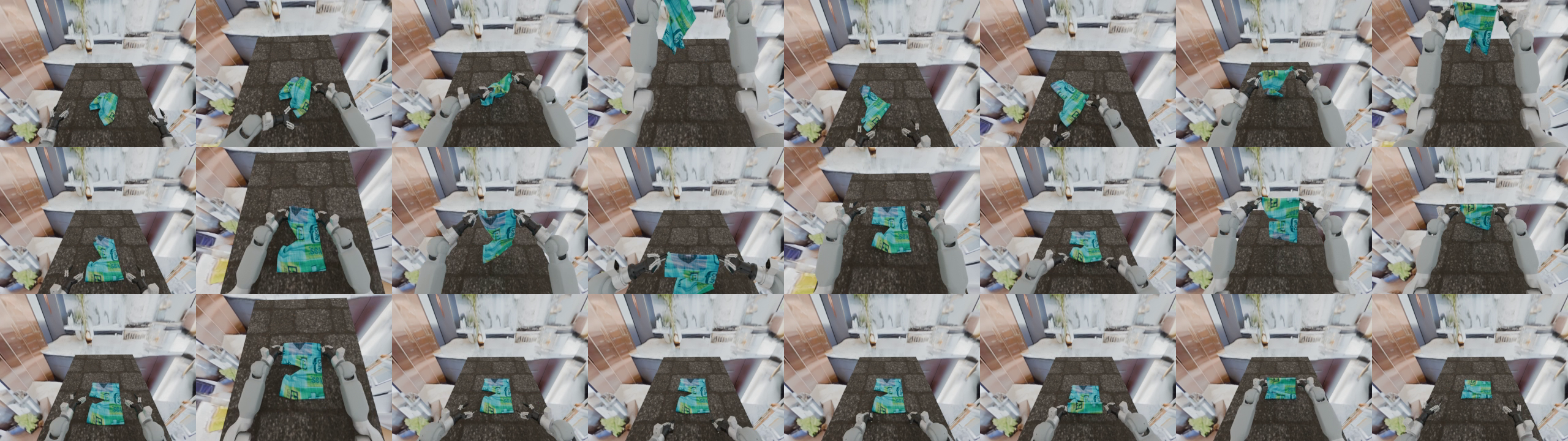}
\caption{Folding and unfolding demonstrations for shorts generated by our pipeline.}
\label{fig:shorts_demo}
\end{figure}

\paragraph{Extension to other garment types.} Extending our pipeline to new garment categories primarily requires designing new category-specific manipulation rules. As a proof of concept, we design folding and unfolding rules for shorts within a single day and successfully generate physical demonstrations in simulation (Fig.~\ref{fig:shorts_demo}). While rule design is fast, achieving strong sim-to-real generalization for a new garment category requires scaling up its corresponding dataset, which we leave for future work.

\end{document}